\documentclass{article}
\usepackage{spconf,amsmath,graphicx}
\usepackage{booktabs}
\usepackage{multirow}
\usepackage{array}   
\usepackage{xcolor}
\usepackage{amssymb}
\usepackage{stmaryrd} 
\usepackage{colortbl}
\usepackage{url}

\usepackage[numbers]{natbib}
\usepackage{enumitem}
\setlist{nosep, leftmargin=14pt}

\usepackage{mwe} 

\title{GMD-Mamba: Lightweight 3D Medical Image Segmentation Using a Grouped Multi-Directional Mamba}

\name{Rahim Ullah$^{1}$, Hasan Abbas$^{2}$, Hazrat Ali$^{3}$ and Syed Waqar Nabi$^{1}$}
\address{$^{1}$School of Computing, University of Glasgow, Glasgow, United Kingdom \\
         $^{2}$James Watt School of Engineering, University of Glasgow, Glasgow, United Kingdom \\
         $^{3}$School of Computing, Data and Mathematical Sciences, University of Stirling, Stirling, United Kingdom \\
         rahim.ullah@glasgow.ac.uk, hasan.abbas@glasgow.ac.uk, ali.hazrat@stir.ac.uk, syed.nabi@glasgow.ac.uk}

\begin{document}
%
\maketitle
\begin{abstract}
Mamba has the capability to capture global context in linear time for 3D medical image segmentation. However, existing methods employ Mamba with full-width directional scans at every resolution, keeping the model large (64–138M parameters and high FLOPs). We propose Grouped Multi-Directional (GMD) Mamba, a lightweight network based on the GMD scan that divides input channels $C$ into groups $G$ and flattens each group in different directions in one pass. This gives multi-directional context at $\mathcal{O}(C^2/G)$ instead of $\mathcal{O}(K \cdot C^2)$, where $K$ is the number of scan directions. Additionally, we employ Mamba only at the deepest encoder stages, resulting in 19.2M parameters. On BraTS2023, it achieves the best average Dice of 92.0\% and the lowest HD95 of 2.7 at ~1/7 the parameters of SegMamba-V2. For model validation, we also ran experiments on the MSD Spleen dataset, achieving competitive performance. The implementation details and code are publicly available at \url{https://github.com/2802990u/GMD-Mamba}.
\end{abstract}

\section{Introduction}
\label{sec:intro}

3D segmentation is a critical task for clinical decisions. CNN architectures have shown strength, however, they cannot capture global context \cite{myronenko20183dmri,cicek20163dunet}. Transformer architecture handle this problem but with quadratic computational cost \cite{devlin2019bert,dosovitskiy2020image}. Mamba architectures offer linear complexity, therefore, have become prominent tool for both 2D and 3D medical image segmentation \cite{xing2025segmambav2,zada2026bisegmamba}. In addition, Mamba has been integrated with SegResNet in \cite{das2025segresmamba} for segmentation. However, the mentioned Mamba-based architectures are computationally heavy, which erode the benefits of State Space Model's (SSM) linear cost. To reduce this cost, group-based scanning has been proposed. GroupMamba~\cite{shaker2025groupmamba} splits channels into groups and scans them in different directions for 2D vision, and GAD-MambaUNet~\cite{gad2026mambaunet} applies direction-group scanning for medical segmentation. However, these methods target 2D images and do not address 3D voxel sequence flattening or multi-directional scanning in volumetric data.

Existing Mamba-based segmentation methods flatten the 3D volume into tokens with a fixed scan order, which exposes each pass to context along only one traversal direction. Capturing 3D spatial relationships therefore requires multi-directional scanning and fusion, as in SegMamba and SegMamba-V2~\cite{xing2024segmamba,xing2025segmambav2}. However, the cost of an SSM layer is dominated by its input and output projections, which scale quadratically with the feature width $C$, so $K$ full-width directional scans cost $\mathcal{O}(KC^2)$. Accordingly, SegMamba has about 67M parameters and SegMamba-V2 has
138.8M, with FLOPs far above those of other state-of-the-art models, which weakens the efficiency argument for SSMs.

To address this limitation, we introduce GMD-Mamba, which integrates a GMD scan to capture 3D spatial context through multi-directional scanning while reducing computational overhead. Instead of running $K$ separate full scans and fusing them, GMD partitions the feature channels into $G$ groups and scans each group along a different direction (standard axis orderings and their reverses) in a single pass, using a compact Mamba of width $\approx C/G$ per group. This reduces the computational cost, $\mathcal{O}(C^2/G)$ instead of $\mathcal{O}(K \cdot C^2)$, with no structure-tensor computation, eigen-decomposition, or learned direction selection. The SSM is integrated into the lower layers of SegResNet’s encoder–decoder, where Mamba does not process more than $16^3$ tokens. Thus, the model has 19.2M parameters. Our contributions are:


\begin{figure*}[t]
\centering
\includegraphics[width=\linewidth, height=9.5cm, keepaspectratio]{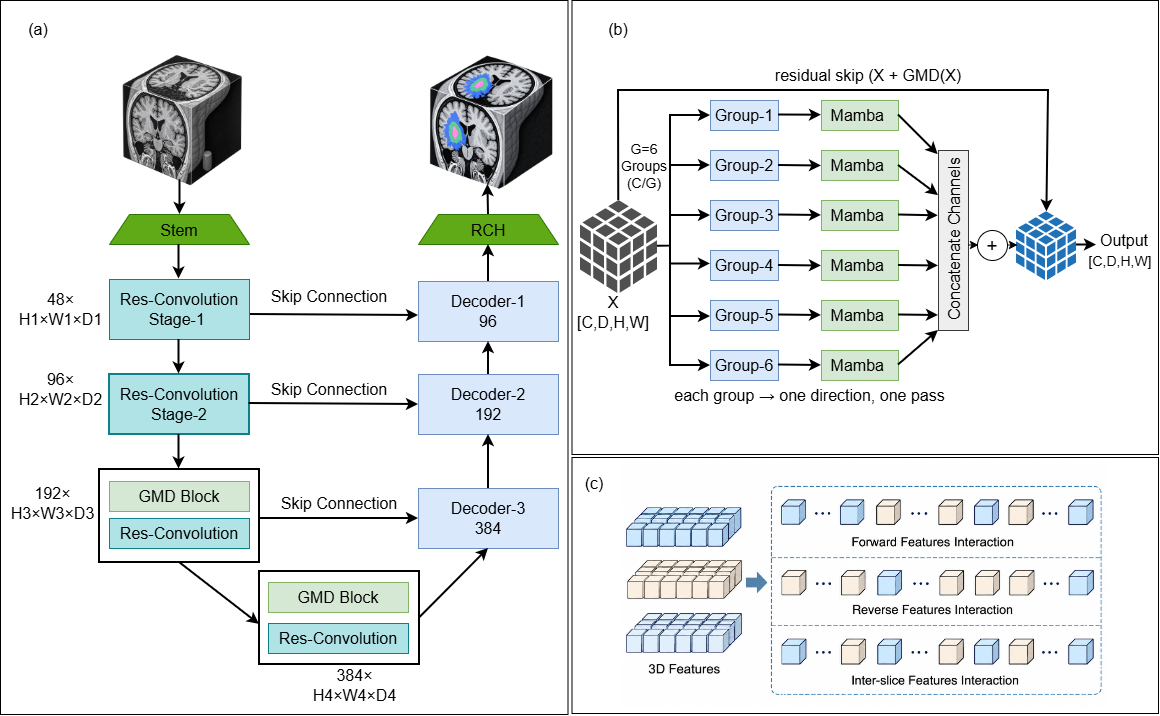}
\caption{Overview of proposed GMD-Mamba: (a) The overall architecture of GMD-Mamba employing GMD block at the lower stage of residual UNet. (b) The GMD block dividing channels in different groups and scanning each group along a different direction. (c) Voxel to sequence flattening for three forward groups.}
\label{fig:framework}
\end{figure*}

\begin{itemize}
\item We propose GMD-Mamba for 3D segmentation, achieving multi-directional context in a single pass by dividing channels into groups with distinct scan orders, reducing complexity from $\mathcal{O}(K \cdot C^2)$ to $\mathcal{O}(C^2/G)$.   
\item We employ GMD at the lower stages of residual UNet only, giving an architecture which is about $7\times$ smaller than SegMamba-V2 while retaining global context.
\item GMD-Mamba achieves the best average Dice of 92.0\% on BraTS2023 and improves Dice by 12.5\% points over its backbone on MSD Spleen dataset.
\item We perform ablation study, which shows that deep-only GMD matched or exceeded the alternative scans.
\end{itemize}

\begin{table*}[t]
\centering
\caption{BraTS2023 comparison under SegMamba-V2’s partition and protocol (baselines as reported in \cite{xing2025segmambav2}). Dice (\%, $\uparrow$) / HD95 (mm, $\downarrow$), mean,$\pm$,std over 250 cases(ET: Enhancing Tumor, TC: Tumor Core, WT: Whole Tumor); best in \textbf{bold}.}
\label{tab:brats}
\setlength{\tabcolsep}{3pt}
\resizebox{\textwidth}{!}{%
\begin{tabular}{c@{\hspace{3pt}}l c ccccccccc}
\toprule
 & & & \multicolumn{2}{c}{WT} & \multicolumn{2}{c}{TC} & \multicolumn{2}{c}{ET} & \multicolumn{2}{c}{Average} \\
\cmidrule(lr){4-5}\cmidrule(lr){6-7}\cmidrule(lr){8-9}\cmidrule(lr){10-11}
Family & Method & Params (M) & Dice & HD95 & Dice & HD95 & Dice & HD95 & Dice & HD95 \\
\midrule
\multirow{4}{*}{\rotatebox[origin=c]{90}{CNN}}
 & SegResNet \cite{myronenko20183dmri}   & 18.80 & 92.02\,\tiny$\pm$6.62 & 4.07\,\tiny$\pm$8.34 & 89.10\,\tiny$\pm$17.00 & 4.08\,\tiny$\pm$7.95 & 83.66\,\tiny$\pm$19.05 & 3.88\,\tiny$\pm$9.25 & 88.26\,\tiny$\pm$14.22 & 4.01\,\tiny$\pm$8.51 \\
 & nnU-Net \cite{isensee2021nnu}     & 31.4  & 93.50\,\tiny$\pm$6.31 & 3.54\,\tiny$\pm$7.58 & 91.20\,\tiny$\pm$14.34 & 3.94\,\tiny$\pm$4.51 & 86.70\,\tiny$\pm$18.15 & 3.55\,\tiny$\pm$8.78 & 90.46\,\tiny$\pm$12.93 & 3.67\,\tiny$\pm$6.95 \\
 & MedNeXt \cite{roy2023mednext}    & 17.6 & 92.41\,\tiny$\pm$7.01 & 4.98\,\tiny$\pm$8.77 & 87.75\,\tiny$\pm$19.19 & 4.67\,\tiny$\pm$6.75 & 83.96\,\tiny$\pm$19.02 & 4.51\,\tiny$\pm$9.90 & 88.04\,\tiny$\pm$15.07 & 4.72\,\tiny$\pm$8.47 \\
\midrule
\multirow{4}{*}{\rotatebox[origin=c]{90}{Transformer}}
 & UNETR \cite{P13}         & 93.0 & 92.19\,\tiny$\pm$9.83 & 6.17\,\tiny$\pm$11.77 & 86.39\,\tiny$\pm$21.60 & 5.29\,\tiny$\pm$11.93 & 84.48\,\tiny$\pm$21.35 & 5.03\,\tiny$\pm$11.82 & 87.68\,\tiny$\pm$17.59 & 5.49\,\tiny$\pm$11.84 \\
 & SwinUNETR \cite{hatamizadeh2022swinunetr}   & 62.2  & 92.71\,\tiny$\pm$7.91 & 5.22\,\tiny$\pm$9.58 & 87.79\,\tiny$\pm$19.95 & 4.42\,\tiny$\pm$7.63 & 84.21\,\tiny$\pm$20.24 & 4.48\,\tiny$\pm$10.17 & 88.23\,\tiny$\pm$16.03 & 4.70\,\tiny$\pm$9.12 \\
 & SwinUNETR-V2 \cite{he2023swinunetrv2}  & 72.8  & 93.35\,\tiny$\pm$6.47 & 5.01\,\tiny$\pm$8.92 & 89.65\,\tiny$\pm$17.14 & 4.41\,\tiny$\pm$7.44 & 85.17\,\tiny$\pm$19.33 & 4.41\,\tiny$\pm$10.56 & 89.39\,\tiny$\pm$14.31 & 4.51\,\tiny$\pm$8.97 \\
 & nnFormer \cite{zhou2023nnformer}     & 149.3 & 91.15\,\tiny$\pm$9.11 & 5.65\,\tiny$\pm$10.91 & 85.94\,\tiny$\pm$19.78 & 5.31\,\tiny$\pm$9.02 & 78.73\,\tiny$\pm$19.66 & 5.09\,\tiny$\pm$10.05 & 85.27\,\tiny$\pm$16.18 & 5.35\,\tiny$\pm$9.99 \\
\midrule
\multirow{4}{*}{\rotatebox[origin=c]{90}{Mamba}}
 & SegMamba \cite{xing2024segmamba}      & 67.4  & 93.61\,\tiny$\pm$6.52 & \textbf{3.37}\,\tiny$\pm$7.78 & 92.65\,\tiny$\pm$12.99 & 3.85\,\tiny$\pm$8.01 & 87.71\,\tiny$\pm$19.23 & 3.48\,\tiny$\pm$10.69 & 91.32\,\tiny$\pm$12.28 & 3.56\,\tiny$\pm$8.66 \\
 & SegMamba-V2 \cite{xing2025segmambav2} & 138.8 & \textbf{94.02}\,\tiny$\pm$6.14 & 3.41\,\tiny$\pm$6.32 & \textbf{92.83}\,\tiny$\pm$11.80 & 2.92\,\tiny$\pm$6.33 & 87.93\,\tiny$\pm$17.57 & 3.36\,\tiny$\pm$8.96 & 91.60\,\tiny$\pm$11.83 & 3.23\,\tiny$\pm$7.20 \\
 & \textbf{GMD-Mamba (ours)} & \textbf{19.20} & 93.89\,\tiny$\pm$6.33 & 3.38\,\tiny$\pm$6.62 & 92.62\,\tiny$\pm$13.21 & \textbf{2.80}\,\tiny$\pm$6.28 & \textbf{89.50}\,\tiny$\pm$14.35 & \textbf{1.87}\,\tiny$\pm$2.69 & \textbf{92.00}\,\tiny$\pm$9.05 & \textbf{2.70}\,\tiny$\pm$3.46 \\
\bottomrule
\end{tabular}%
}
\end{table*}
\begin{table}[hb!]
\centering
\caption{Ablation of the scan design on BraTS2023 (same
partition; region-wise average Dice, $400$ epoch). }
\label{tab:ablation}
\begin{tabular}{lcc}
\toprule
Configuration & Params (M) & Dice (\%) \\
\midrule
Backbone, no Mamba              & 19.10 & 91.31 \\
\;+ Bidirectional scan (deep)   & 19.65 & 91.17 \\
\;+ GMD scan (shallow+deep)     & 19.24 & 89.85 \\
\;+ GMD scan (deep) \textbf{[ours]} & \textbf{19.23} & \textbf{91.47} \\
\bottomrule
\end{tabular}
\end{table}

\begin{table}[hb!]
\centering
\caption{Segmentation Results on MSD~Spleen}
\label{tab:spleen}
\begin{tabular}{llcc}
\toprule
 & Method & Params (M) & Dice (\%) \\
\midrule
 & SegResNet (backbone) & 18.80 & 79.4 \\
 & \textbf{GMD-Mamba}    & \textbf{19.20} & \textbf{91.9\,$\pm$\,2.5} \\
\midrule
 & 3D U-Net \cite{cicek20163dunet}     & 4.9   & 81.95 \\
 & UNETR \cite{P13}         & 93.0  & 86.42 \\
 & SegMamba \cite{xing2024segmamba}      & 67.0  & 90.04 \\
 & SwinUNETR \cite{hatamizadeh2022swinunetr}     & 62.0  & 91.26 \\
\bottomrule
\end{tabular}\\[4pt]
\raggedright \footnotesize \textit{Note:} All models are trained and evaluated by us under the same protocol.
\end{table}

\section{Methodology}
\label{sec:method}

Our proposed designed model (see Fig. ~\ref{fig:framework}) is based on a single principle that convolution, especially SegResNet, is efficient and effective for local structures at high resolution but cannot capture long range dependencies. That is why convolution is used at high resolution stages while Mamba is employed only at lower stages where token count is small and long range context is most useful.

The backbone of the model is a SegResNet-style U-Net. First, a $3\times3\times3$ stem takes the input to $C$ channels, and each stage uses pre-activation residual blocks (GroupNorm$\rightarrow$ReLU$\rightarrow$$3\times3\times3$ convolution, applied twice with an additive shortcut). The encoder consists of four stages with channel widths $(C,2C,4C,8C)$ and reduces the spatial resolution by half between successive stages using a stride-2 convolution; the decoder mirrors this architecture by up-sampling the features accordingly. Finally, a reconstruction head (RCH) adds the matching encoder feature to the decoder feature, refines the fused feature map with a residual block, and passes the result to a $1\times1\times1$ segmentation head that outputs the class logits. For a $128^3$ input, the four stages operate at the resolution of $64^3$, $32^3$, $16^3$ and $8^3$.

We apply Mamba~\cite{gu2023mamba} to 3D segmentation by flattening the feature map into a token sequence and processing it with a SSM block. The computational cost of SSM is dominated by the input projections, which scale as $\mathcal{O}(C^2)$ in the feature width $C$. To feed a 3D feature map to such a model, it needs to be converted to a sequence, but this single flattening order captures context in only one traversal order. Capturing the actual 3D relationships requires $K$ full-width directional scans that are subsequently fused, at a cost of $\mathcal{O}(K.C^2)$.

GMD captures the multi-directional coverage in a single pass by partitioning channels across directions as shown in Fig.~\ref{fig:framework}(b). Given a feature map $X\in\mathbb{R}^{C\times D\times H\times W}$, we flatten it into 1D sequence tokens, apply LayerNorm, then split the $C$ channels into $G$ groups. Each group $g$ is assigned a distinct scan order $\pi_g$ drawn from the axis permutations and their reversals $\{(D,H,W),(H,D,W),(W,D,H)\}\times\{\text{forward},\text{reverse}\}$, and is processed by a lightweight Mamba operator with a width of $\approx\! C/G$. To restore the original voxel order, the group outputs are concatenated and added residually.
\begin{equation}
\mathrm{GMD}(X) = X + \bigparallel_{g=1}^{G}
\pi_g^{-1}\big(\mathrm{Mamba}_g(\pi_g(X_g))\big),
\label{eq:gmd}
\end{equation}
where $\bigparallel$ denotes channel concatenation, $\pi_g$ and $\pi_g^{-1}$ denote the scan-order permutation and its inverse for the $g$th group respectively, and $X_g$ represents the $g$th channel group. Since each group processes only $C/G$ channels, the corresponding projection cost is reduced to
\begin{equation}
G\cdot\mathcal{O}\big((C/G)^2\big) = \mathcal{O}(C^2/G),
\label{eq:cost}
\end{equation}
an approximately $KG$-fold reduction relative to $K$ full-width scans. Besides, GMD does not use any structure tensor estimation, eigen decomposition or learned direction selection, which keeps the mixer computationally efficient and simple.

We keep the GMD mixer only at the two deepest encoder stages where resolution is $16^3$ and $8^3$ (for $16^3\!=\!4096$ tokens), leaving the higher stages purely convolutional, where the token count would otherwise reach hundreds of thousands per scan. The ablation study (Table~\ref{tab:ablation}) shows that keeping the mixer at earlier  stages does not improve accuracy, and that naive full-width multi-directional scans can even degrade the convolutional backbone at this scale.

\section{Results and Discussion}
\label{sec:results}

\noindent\textit{Datasets and protocol.}
On BraTS2023, we kept SegMamba-V2’s published partition and training recipe (SGD, lr $10^{-2}$, momentum $0.99$, cross-entropy loss, $1000$ epochs, batch size $2$, $128^3$ patches); the only change is the network architecture, so our results are directly comparable to theirs. The one deviation is that we select checkpoints on a held-out validation set instead of the test set, which makes our numbers conservative but avoids test-set leakage and overfitting and ensures unbiased evaluation. On MSD~Spleen, our model is trained with nnU-Net-style pre-processing, Dice-cross-entropy loss and Adam. We report Dice on a held-out test set.

\noindent\textit{Comparison with the state of the art (BraTS2023).}
Table~\ref{tab:brats} compares GMD-Mamba with SOTA CNN, Transformer and Mamba methods on the same dataset, partition and protocol. Our model achieves the highest average Dice of 92.0\% and ET score while using only $19.2$M parameters, roughly $7\times$ fewer than SegMamba-V2 ($138.8$M) and lighter than most of the models. The strongest competitor, SegMamba-V2, is marginally higher on WT and TC but is $7\times$ larger. Thus, the proposed model has the highest average Dice and lowest average HD95. 

\noindent\textit{Ablation study.}
Table~\ref{tab:ablation} shows the ablation study on the effectiveness of GMD at different stages and positions at $400$ epochs, same partition and protocol (rows are comparable to each other, not the $1000$-epoch headline). Adding the GMD scan to the deep stage results in improving the dice score on BRaTS2023 ($91.31\!\rightarrow\!91.47$) at a negligible parameter cost. The plain bidirectional scan and a full-width multi-directional scan increase the parameters and yet result in a lower accuracy.  Furthermore, extending GMD to high resolution stages reduces accuracy ($91.47\!\rightarrow\!89.85$) without any reduction in the parameters count, supporting the deep-only placement. A gated fusion variant yielded no accuracy gains.

\noindent\textit{Spleen segmentation.}
The results of GMD-Mamba on the Spleen benchmark are shown in Table~\ref{tab:spleen}. It achieved a competitive performance with a Dice score of 91.9\%, which is 12.5 percentage points higher than SegResNet (79.4\%), with only 0.4M additional parameters. GMD-Mamba outperformed all baseline models in Dice score despite having approximately four times fewer parameters than most of them, except 3D U-Net. This suggests that GMD-Mamba offers benefits on both the segmentation tasks, particularly Spleen segmentation, where the convolutional backbone alone performs poorly.

\begin{figure}[t]
  \centering
  \includegraphics[width=\linewidth, height=4.5cm]{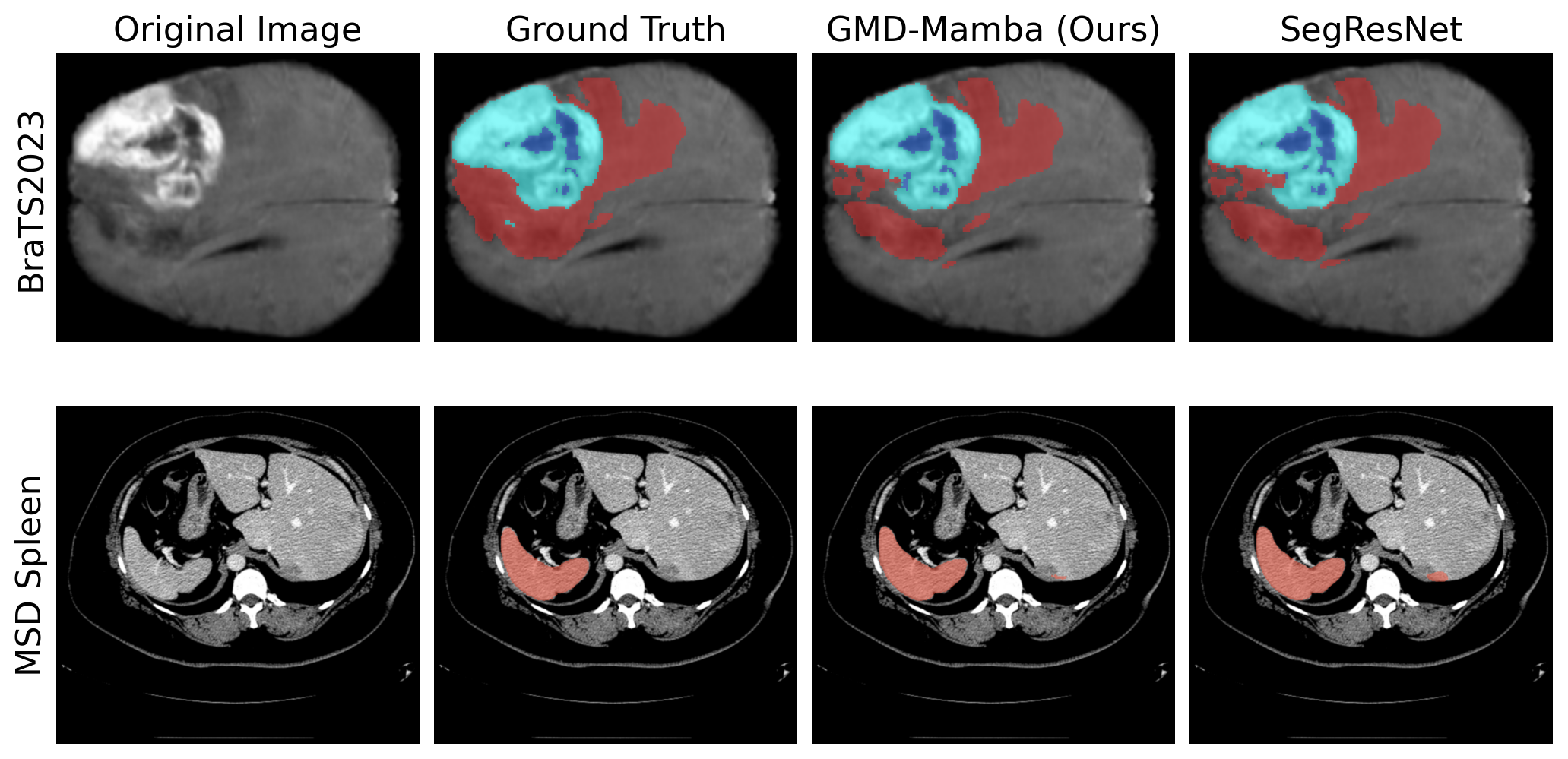}
  \caption{Qualitative results on BraTS2023 (top) and MSD~Spleen (bottom):
  input, ground truth, GMD-Mamba (ours), and SegResNet's predicted.}
  \label{fig:qual}
\end{figure}

\noindent\textit{Qualitative results.}
Fig.~\ref{fig:qual} shows axial slices of the input, ground truth, GMD-Mamba, and SegResNet predictions for representative BraTS2023 and spleen cases. The predicted masks match closely with the GT, including at TC and ET boundaries, consistent with the quantitative gains in ET.

\section{Conclusion}
\label{sec:discussion}
We presented GMD-Mamba, a lightweight 3D segmentation network that achieves multi-directional state-space context at $\mathcal{O}(C^2/G)$ scan cost. With $19.2$M parameters, it achieves the best average Dice among the compared methods on BraTS2023, using approximately $1/7$ the parameters of SegMamba-V2, and improves performance over its convolutional backbone on MSD Spleen. However, the small MSD Spleen test set and evaluation on only two datasets limit generalizability of the method. Future work will evaluate the method across additional anatomies and imaging modalities.

\section{Compliance with Ethical Standards}
This study utilized publicly available benchmark datasets. Ethical approval was not required, as confirmed by the licenses attached to the open-access data.

\section{Acknowledgments}
This work has been supported by funding from Higher Education Commission (HEC), Pakistan under the ASDIS through international scholarships scheme. The authors declare no conflicts of interest. 

\small
\bibliographystyle{IEEEbib}
\bibliography{strings,refs}

\end{document}